\documentclass[letterpaper, 10 pt, conference]{ieeeconf}  

\IEEEoverridecommandlockouts                              

\usepackage{graphicx}
\usepackage{cite}
\usepackage{amsmath,amssymb,amsfonts}
\usepackage{algorithm}
\usepackage{algorithmic}
\usepackage{comment}
\usepackage{caption}
\usepackage{subcaption}
\usepackage{multirow}
\usepackage{float}    
\usepackage{placeins} 
\usepackage{hyperref}
\usepackage{booktabs}
\usepackage{svg}

\usepackage{eso-pic}
\newcommand{\mycopyrighttext}{%
  \footnotesize
  \noindent
  \textcopyright~2025 IEEE. Personal use of this material is permitted. Permission from IEEE must be obtained for all other uses, in any current or future media, including reprinting/republishing this material for advertising or promotional purposes, creating new collective works, for resale or redistribution to servers or lists, or reuse of any copyrighted component of this work in other works.\\
  IEEE 36th Intelligent Vehicles Symposium (IV 2025) - 22-25 June, 2025.
}

\AtBeginDocument{%
  \AddToShipoutPicture*{%
    \AtPageUpperLeft{%
      \put(55, -40){
        \parbox{\dimexpr\textwidth-4pt\relax}{\raggedright \mycopyrighttext}
      }%
    }
  }
}

\title{\LARGE \bf
Automatic Curriculum Learning for Driving Scenarios: Towards \\ Robust and Efficient Reinforcement Learning}
\author{Ahmed Abouelazm$^{1}$, Tim Weinstein$^{2}$, Tim Joseph$^{1}$, Philip Schörner$^{1}$, and J. Marius Zöllner$^{1,2}$
\thanks{$^{1}$Authors are with the FZI Research Center for Information Technology, Germany
        {\tt\small abouelazm@fzi.de}}%
\thanks{$^{2}$Authors are with the Karlsruhe Institute of Technology, Germany}%
}

\begin{document}
\maketitle
\thispagestyle{empty}
\pagestyle{empty}

\begin{abstract}
    This paper addresses the challenges of training end-to-end autonomous driving agents using Reinforcement Learning (RL). RL agents are typically trained in a fixed set of scenarios and nominal behavior of surrounding road users in simulations, limiting their generalization and real-life deployment. While Domain Randomization offers a potential solution by randomly sampling driving scenarios, it frequently results in inefficient training and sub-optimal policies due to the high variance among training scenarios. To address these limitations, we propose an automatic curriculum learning framework that dynamically generates driving scenarios with adaptive complexity based on the agent's evolving capabilities. Unlike manually designed curricula that introduce expert bias and lack scalability, our framework incorporates a "teacher" that automatically generates and mutates driving scenarios based on their learning potential—an agent-centric metric derived from the agent's current policy, eliminating the need for expert design. The framework enhances training efficiency by excluding scenarios the agent has mastered or finds too challenging. We evaluate our framework in a reinforcement learning setting where the agent learns a driving policy from camera images. Comparative results against baseline methods, including fixed scenario training and domain randomization, demonstrate that our approach leads to enhanced generalization, achieving higher success rates, +9\% in low traffic density, +21\% in high traffic density, and faster convergence with fewer training steps. Our findings highlight the potential of ACL in improving the robustness and efficiency of RL-based autonomous driving agents.
    
\end{abstract}
\section{Introduction}
\label{sec:Introduction}
End-to-end (E2E) driving has emerged as a reliable approach in the field of autonomous vehicles, shifting away from traditional modular systems that rely on predefined sub-tasks for perception, planning, and control~\cite{endtoendreview}. Instead, E2E approaches leverage neural networks to directly map raw sensor data to driving actions~\cite{wang2021advsim}. By allowing the neural network to develop an implicit understanding of the driving environment, E2E driving offers a more flexible and adaptive solution for automated driving tasks~\cite{endtoendreviewpaper}.

One promising approach for achieving E2E driving is Reinforcement Learning (RL). In RL, the agent learns by interacting directly with the environment, guided by a reward function~\cite{Sutton1998}. By maximizing a cumulative reward, RL agents can iteratively improve their decision-making policies and ultimately learn to navigate complex environments~\cite{kiran2021deep}. RL-based E2E approaches have demonstrated the ability to handle various input modalities, including object-level information~\cite{naveed2021trajectory}, images~\cite{bogdoll2024informed}, LiDAR~\cite{mammadov2023end}, and multimodal inputs~\cite{chen2021interpretable}. Furthermore, RL approaches have demonstrated success across different decision-making levels, from high-level behavioral planning~\cite{caoHighwayExitingPlanner2021} over trajectory planning~\cite{naveed2021trajectory} to direct control~\cite{wu2022uncertainty}.

\textbf{Research Gap.}
RL driving agents are typically trained in simulations to safely explore various driving commands in numerous interactions with the environment. However, previous works~\cite{chen2019model} often restrict training to a fixed set of driving scenarios with nominal behavior from surrounding road users, such as all vehicles driving at a constant velocity. This approach simply leads to overfitting the agent's policy on the training scenarios~\cite{gisslen2021adversarial}, as it fails to expose agents to diverse scenarios with different road layouts and varying numbers and behaviors of road users. Consequently, RL agents struggle to develop robust policies that effectively generalize to unseen driving scenarios.

To mitigate this limitation, researchers have explored diverse scenario-generation methods. Domain Randomization (DR)~\cite{tobin2017domain} introduces variability by randomly generating scenarios during training, enabling agents to learn generalizable driving behaviors rather than memorizing specific details of the environment. While DR improves performance and generalization, it suffers from low sample efficiency~\cite{josifovski2022analysis} and often results in suboptimal policies due to high variance in training scenarios~\cite{hansen2020self}.

A more advanced approach is Curriculum Learning (CL)~\cite{matiisen2019teacher}, where, unlike DR, tasks are presented to the agent in a structured sequence, starting with simpler tasks and gradually progressing to more complex ones to enhance learning and generalization. This approach mimics the way humans learn, with a “teacher” selecting tasks suited to the current capabilities of the “student” (the agent). Most Autonomous Driving (AD) approaches that utilize CL~\cite{khaitan2022state, anzalone2021reinforced} rely on manually designed curricula, where driving scenarios are divided into discrete stages of increasing difficulty based on expert-defined heuristics. Additional heuristics are necessary to manage the agent’s progression between these stages. 
While effective for training, these approaches are labor-intensive and introduce human bias, making adaptation to new environments challenging.

\textbf{Contribution.}
This work presents a novel framework for the automatic generation of diverse driving scenario curricula, overcoming key limitations of existing RL training frameworks—namely, training inefficiency and the reliance on manually designed curricula with expert-defined heuristics. The key contributions of this work are:
\begin{itemize}
    \item \textbf{Graph-Based Environment Representation}: A flexible graph-based representation of driving environments where nodes and edges represent free parameters that can be dynamically modified to generate diverse and challenging new scenarios. 
    
    \item \textbf{Automatic Scenario Generation}: A teacher-student framework where a teacher dynamically generates new scenarios or mutates existing ones based on the student policy’s evolving capabilities. Thus, this approach eliminates the need for manual curriculum design and ensures a gradual progression in complexity.
    
    \item \textbf{Framework Evaluation}: We conduct a thorough evaluation of the proposed framework, analyzing its impact on training efficiency, policy generalization, and the progression of scenario complexity.
\end{itemize}
\begin{figure}[t!]
    \centering
    \begin{subfigure}[t]{\linewidth}
    \centering
    \includegraphics[width=\linewidth]{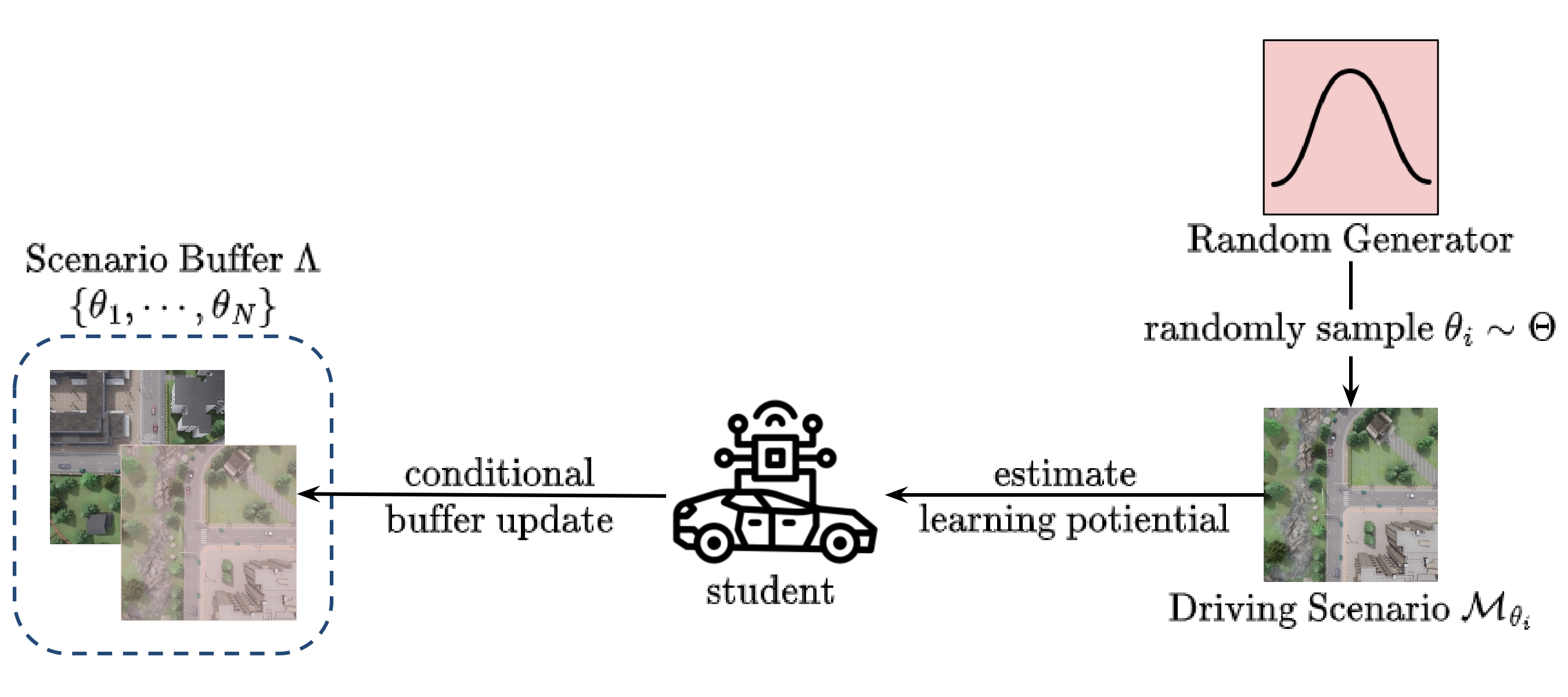}
    \caption{Diagram of the proposed algorithm for sampling new scenarios randomly (replay decision $d = 0$).}
    \label{fig:algorithm_random}
    \end{subfigure}%
    \vfill
    \vspace{0.3cm}
    \begin{subfigure}[t]{\linewidth}
    \centering
    \includegraphics[width=\linewidth]{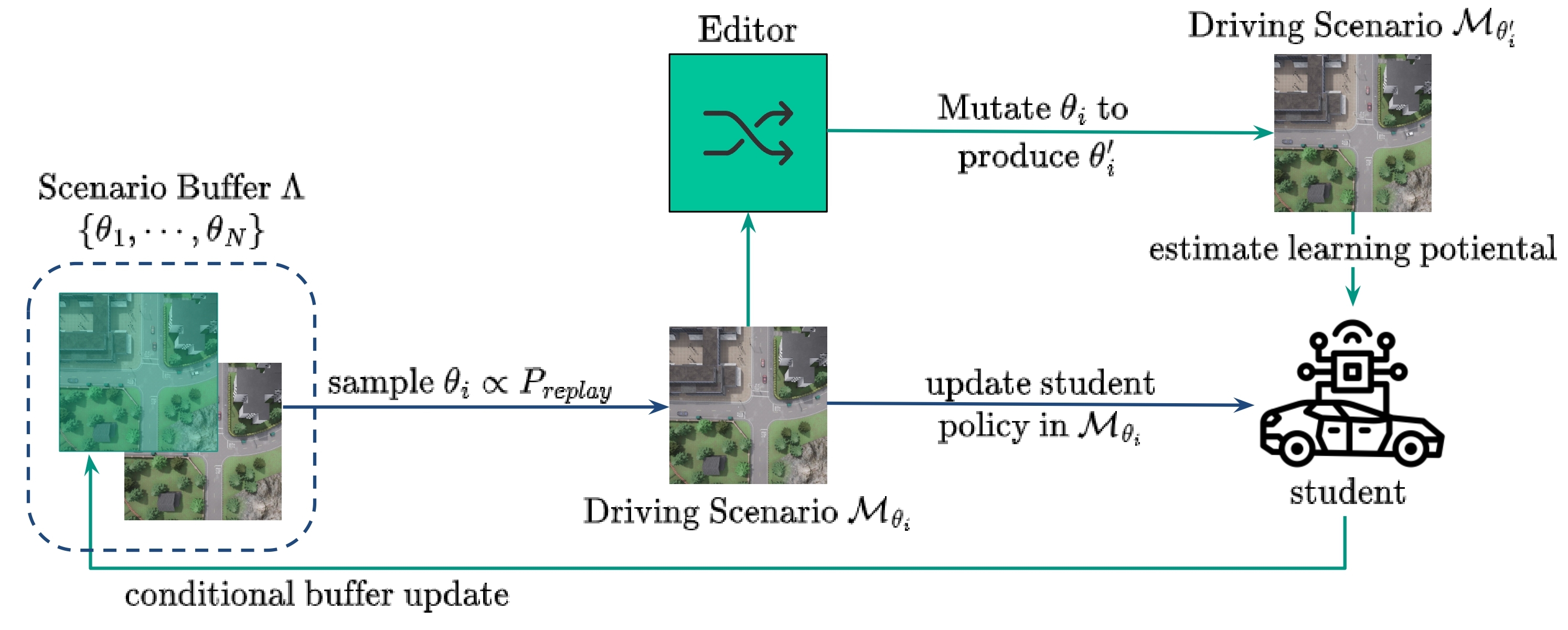}
    \caption{Diagram of the proposed algorithm when the student policy is updated using scenarios sampled from the scenario buffer and further mutated via the editor (replay decision $d = 1$).}
    \label{fig:algorithm_edit}
    \end{subfigure}
    \caption{The proposed framework alternates between two modes based on the replay decision $d$. When $d = 0$, a random generator creates diverse scenarios by varying environment parameters. Scenarios with high learning potential are conditionally added to the scenario buffer $\Lambda$, ensuring training efficiency. When $d = 1$, the student trains exclusively on scenarios sampled from $\Lambda$, while an editor mutates them to further refine the most effective scenarios for enhancing the student's learning progress.}
    \label{fig:accel_overall_framework}
    \vspace{-0.2cm}
\end{figure}

\section{related work}
In this section, we review related work on various training frameworks for AD and the application of CL. Additionally, we discuss the recent advancement in automatic curriculum learning (ACL) for RL.
\subsection{Training Frameworks for RL in AD}
Training RL agents for AD has traditionally relied on a \textbf{fixed set} of scenarios~\cite{yu2016deep,chen2019model,ashwin2023deep} and \textbf{DR}~\cite{pouyanfar2019roads,kontes2020high,candela2022transferring}. However, fixed training frameworks restrict the agent’s robustness and ability to generalize. In contrast, training with DR, which relies on randomly generated scenarios, can improve generalization but tends to be sample inefficient, requiring longer training and often leading to sub-optimal policies~\cite{josifovski2022analysis,hansen2020self}.

Alternatively, \textbf{CL} has been explored in AD to enhance the training of RL agents and has shown a positive impact on learned policies. A multi-stage CL approach with progressively challenging driving scenarios is proposed in~\cite{anzalone2021reinforced,anzalone2022end}. Each stage increases the difficulty by increasing the number of actors in the scenario, such as pedestrians and vehicles, and varying the RL agent’s starting position. The teacher determines whether to increment the difficulty level based on expert-defined heuristics, such as the percentage of successful episodes. Another curriculum, proposed in~\cite{khaitan2022state}, gradually omits a subset of the student’s state observation, such as the positions and velocities of other actors. The omitted subset size increases as training progresses at a predefined number of episodes. Notably, the proposed curricula are manually designed and rely on rule-based heuristics to increase training difficulty, limiting their scalability.

To the best of our knowledge, only one \textbf{ACL} approach has been proposed in the context of AD~\cite{qiao2018automatically}. However, this approach is insufficient, involving only a single vehicle alongside the RL agent and merely adjusting its starting position. Furthermore, it employs a tabular RL algorithm to learn the curriculum, which does not scale for diverse scenario generation, resulting in insufficiently robust agents. Thus, a scalable ACL approach capable of handling diverse road topologies and actor configurations remains an open challenge in AD.
\subsection{Recent Advancements in ACL}
Recent work on Unsupervised Environment Design (UED)~\cite{dennis2020emergent} has advanced approaches for ACL. UED relies on environments with free parameters to generate diverse levels, where each level represents an environment instance configured with specific parameter values. Adversarial Domain Randomization (ADR)~\cite{khirodkar2018adversarial} is one of the first UED instances. It formulates a two-player game between a student and a teacher, where the teacher is trained to select environment parameters that minimize the student’s performance. While ADR demonstrated that adversarially generated environments can accelerate learning, it suffers from instability and can produce unsolvable environments~\cite{dennis2020emergent} due to the lack of mechanisms to balance between difficulty and learnability.

PAIRED~\cite{dennis2020emergent} addresses this training instability by introducing an antagonist agent to regularize the teacher's policy. The teacher generates levels that maximize the difference in discounted returns between the antagonist and the student. The antagonist acts as a regularizer: environments where the antagonist succeeds but the student fails are favored, ensuring that generated tasks remain solvable yet challenging. This results in a more stable curriculum. However, PAIRED is computationally expensive and relies on multi-agent RL, which lacks convergence guarantees. 

More efficient approaches, such as those in~\cite{jiang2021prioritized, jiang2021replay}, shift the focus from learned teachers to heuristic-driven, sampling-based teachers. In Prioritized Level Replay (PLR)~\cite{jiang2021replay}, a buffer is populated with levels sampled randomly from the environment’s parameter space. The student then trains on a subset of levels selected from this buffer based on their estimated learning potential. While PLR outperforms earlier UED methods in terms of stability and generalization, it cannot modify existing levels to refine the curriculum further. To overcome this limitation, ACCEL~\cite{parker2022evolving} builds on PLR by introducing an editor that mutates high-potential levels. This allows the curriculum to evolve more dynamically, enabling smoother progression and further improvements in the policy.

\section{Methodology}
This section outlines the proposed methodology, as shown in Fig.~\ref{fig:accel_overall_framework}. We first introduce the essential background of our framework, followed by a description of the driving scenario representation and how the teacher utilizes it to construct the curriculum. Finally, we present the underlying algorithm.
\subsection{Background}
\subsubsection{\textbf{Underspecified Partially Observable MDP}} 

Our proposed framework builds on ACCEL~\cite{parker2022evolving}, which extends the traditional Markov Decision Process (MDP)~\cite{sutton1998reinforcement} into an Underspecified Partially Observable MDP (UPOMDP)~\cite{dennis2020emergent}. The UPOMDP formulation is presented in Eq.~\ref{eq:upomdp}, where $\mathcal{M_{\text{POMDP}}}$ denotes the POMDP introduced in~\cite{kaelbling1998planning}, and $\boldsymbol{\Theta}$ represents the environment’s free parameters.
\begin{equation}
   \mathcal{M} = \mathcal{M_{\text{POMDP}}} \,+ \,\left\langle \boldsymbol{\Theta} \right\rangle =  \left\langle A, \mathcal{O}, \boldsymbol{\Theta} , \mathcal{S}, \mathcal{T}, \mathcal{I}, \mathcal{R}, \gamma\right\rangle
   \label{eq:upomdp}
\end{equation}
By assigning specific values to the free parameters $\theta \in \boldsymbol{\Theta}$, diverse levels $\mathcal{M}_{\theta}$ are generated, each defining a unique environment instance. In AD, a level corresponds to a driving scenario with varying configurations. Moreover, a POMDP variant is well-suited for designing such scenarios, as a realistic student inherently relies on partial observations of the environment state through sensors.
\subsubsection{\textbf{Unsupervised Environment Design}}
In UED~\cite{dennis2020emergent}, a teacher constructs a sequence of levels by setting the environment's free parameters $\boldsymbol{\Theta}$ of a UPOMDP, guided by a utility function $U(\pi, \theta)$ that adapts to the current student policy $\pi$. A simple example of a UED approach is DR, where the utility function is a constant and independent of the student’s policy. A more advanced class of utility functions based on learning potential is introduced in~\cite{dennis2020emergent}. These functions guide the teacher to generate levels that maximize the gap between the expected return of the student’s policy and the optimal policy. Consequently, these functions have been proven to drive the teacher to construct the simplest levels that the student cannot yet solve~\cite{jiang2021prioritized, jiang2021replay, parker2022evolving}, fostering a curriculum that gradually increases in complexity as the student’s learning progresses. Since the optimal policy is typically unknown, the learning potential utility function is approximated in practice~\cite{parker2022evolving}.
\begin{figure}[t!]
    \centering
    \begin{subfigure}[t]{0.48\linewidth}
    \centering
    \includegraphics[height=0.85\linewidth]{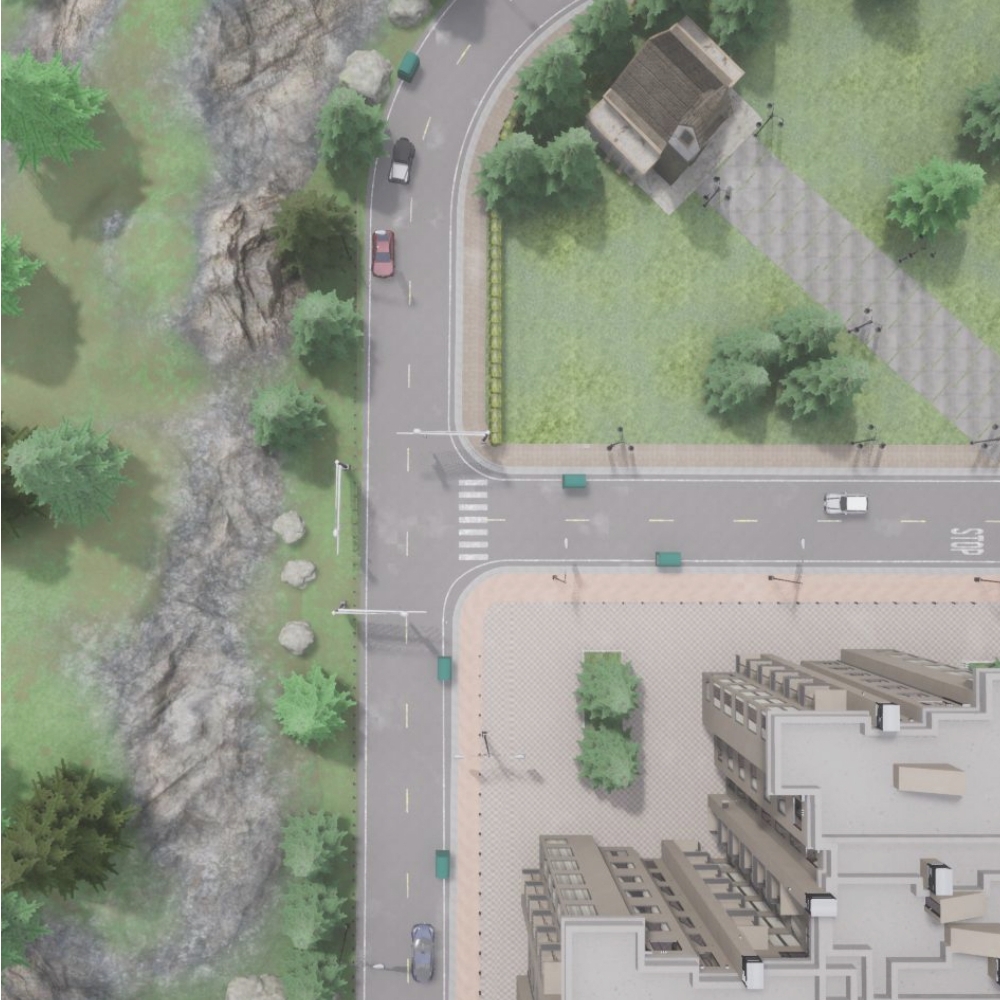}
    \caption{Bird's Eye View perspective of an exemplary driving scenario.}
    \end{subfigure}%
    \hfill
    \begin{subfigure}[t]{0.48\linewidth}
    \centering
    \includegraphics[width=\linewidth]{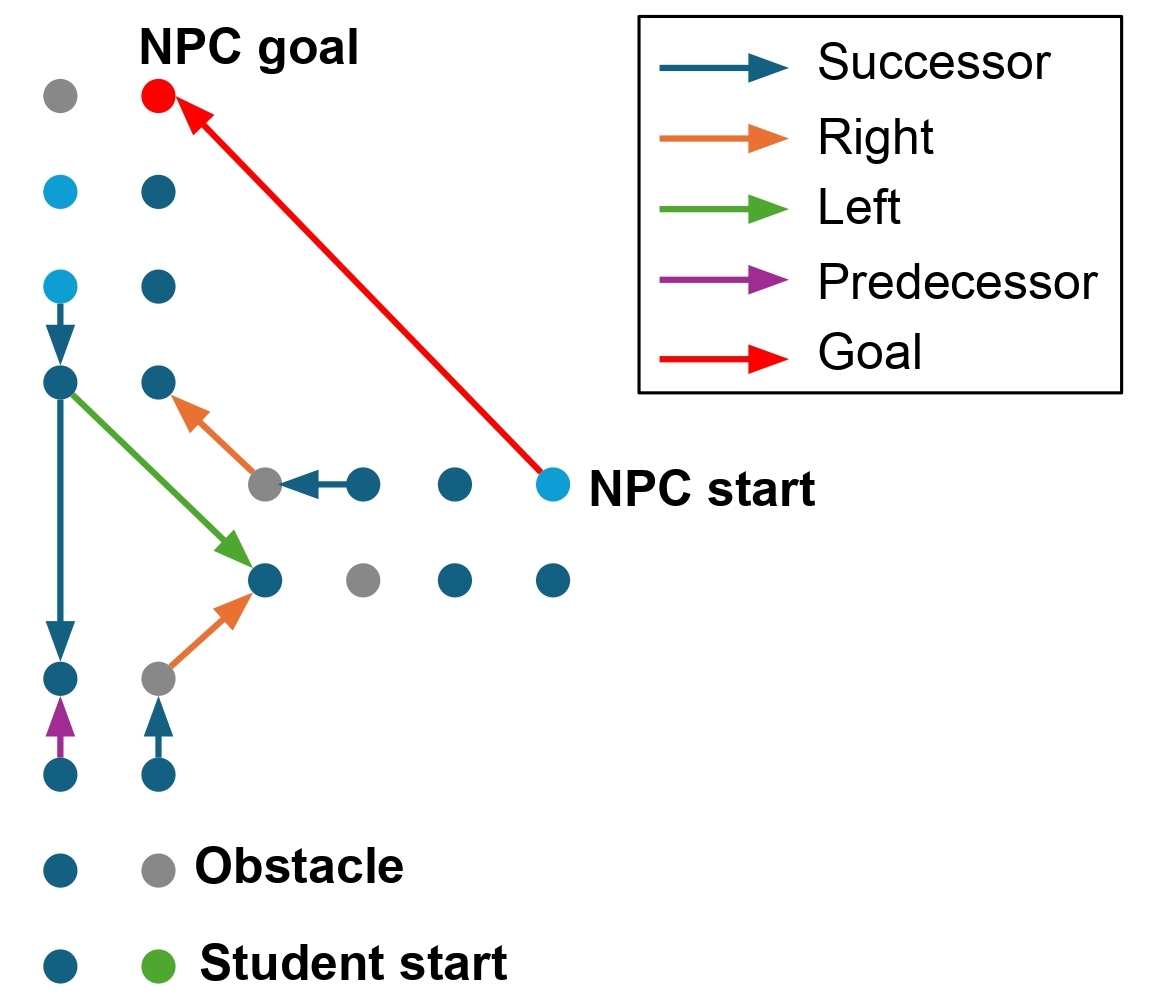}
    \caption{Graph-based representation of an exemplary driving scenario.}
    \label{fig:environment_representation_2}
    \end{subfigure}
    \caption{A driving scenario represented as a directed graph, visualized from a Bird’s Eye View perspective. Nodes are sampled at equidistant intervals along the road and can be occupied by the student, NPCs, obstacles, or remain empty. Edges define the road topology and the goal destinations for the student and NPCs.}
    \label{fig:environment_representation}
    \vspace{-0.2cm}
\end{figure}
\subsection{Framework Overview}
The proposed framework automatically designs a driving scenario curriculum, as shown in Fig.~\ref{fig:accel_overall_framework}. First, we represent the driving environment as a UPOMDP by defining its free parameters. To facilitate curriculum generation, the framework employs a teacher, which consists of two components: a random generator and an editor. The random generator synthesizes scenarios by randomly sampling free parameters, while the editor mutates existing scenarios.

To integrate these components into the training process, our ACL algorithm alternates between two distinct phases. One phase trains the student on scenarios sampled from the scenario buffer $\Lambda$, which has a maximum size of $N$, followed by editing to refine the most effective scenarios for enhancing the student’s learning progress. Another phase involves sampling new scenarios using the random generator to maintain curriculum diversity. To ensure efficient training, only scenarios with high learning potential, whether edited or randomly sampled, are added to $\Lambda$, as low-potential scenarios do not contribute to the student’s learning progress. The following sections discuss framework components in detail.
\subsection{Environment Representation}
While images are commonly used to represent driving environments~\cite{chitta2024sledge,suo2021trafficsim}, they are less suitable for scenario generation due to their dense and rotation-variant nature, which complicates actor placement in feasible areas and requires cumbersome masking. Instead, we propose a graph-based representation to model the driving environment as a UPOMDP, where nodes and edges represent the environment's free parameters $\Theta$. Since graphs provide a sparse representation that facilitates feasible actor placement and effectively captures road topology through edges~\cite{abeysirigoonawardena2019generating}.

We represent a driving environment as a directed graph $G = \left( V, E \right)$, where $V$ denotes the nodes (vertices) and $E$ represents the edges. Nodes are sampled at equidistant intervals along the road topology, as illustrated in Fig.~\ref{fig:environment_representation_2}. Each node can be occupied by the student, a Non-Player Character (NPC), an obstacle, or remain empty. Restricting nodes to valid road positions~\cite{tan2021scenegen,feng2023trafficgen} ensures that all generated scenarios remain feasible by enabling a proper placement of road-bound NPCs, such as vehicles and motorcycles, as well as roadside obstacles like parked cars or trash containers, without requiring additional masking. Roadside obstacles are explicitly included, as they require NPCs to perform avoidance maneuvers, making scenarios more interactive. In addition to node placement, the connectivity between nodes plays a crucial role in defining the driving environment.

Each edge in the graph is assigned a discrete type to represent specific road relationships between nodes, such as successor, predecessor, left, and right~\cite{liang2020learning,vivekanandan2024ki}. These edges are crucial for the teacher to accurately sample goal destinations for the student and NPCs through reachability analysis. When a destination is selected, a goal edge is created with two configurable attributes for NPCs: desired velocity and offset from the centerline. These attributes enhance behavioral diversity among NPCs, contributing to a more dynamic curriculum. To enhance the flexibility of our representation, we encode the spatial information of nodes using rotation- and translation-invariant edge attributes, as proposed in~\cite{cui2023gorela}. This invariant encoding is particularly advantageous for a deep learning-based editor.

In summary, within the proposed environment representation, the free parameters $\boldsymbol{\Theta}$ of the UPOMDP include the actor type assigned to each node, goal destinations for the student and NPCs, as well as the desired velocity and offset, which shape NPC behavior.
\subsection{Teacher-Guided Scenario Design}

To effectively leverage these free parameters in scenario generation, our framework employs a teacher composed of two integral components: a random generator and an editor. These components work in tandem to ensure a dynamic and effective curriculum that adapts to the student's evolving capabilities.

\subsubsection{\textbf{Random Generator}} This generator maintains curriculum diversity by exploring the free parameter space $\boldsymbol{\Theta}$ and randomly sampling unseen scenarios $\mathcal{M}_{\theta}$. The generator employs an autoregressive process~\cite{feng2023trafficgen} to construct new scenarios, described by the following steps:
\begin{enumerate}
    
    \item \textbf{Sample driving environment}: Select a road layout from the available set and initialize it as an empty driving environment. This step ensures a diverse range of road topologies in the curriculum.
    
    \item \textbf{Initial scenario generation}: Assign the student to a start node and sample a valid goal destination. This step defines the student’s primary task and movement within the scenario.
    
    \item \textbf{Actors sampling}: Determine the number of actors, their classes, and their assigned nodes. This ensures that the scenario includes a realistic distribution of classes to create varied and challenging interactions.
    
    \item \textbf{NPC configuration}: Define goal destinations for sampled NPCs and set their desired velocity and offset.
\end{enumerate}

\subsubsection{\textbf{Editor}} In contrast, the editor mutates scenarios $\mathcal{M}_{\theta}$ sampled from the scenario buffer $\Lambda$ to generate variations $\mathcal{M}_{\theta'}$ with equal or slightly higher learning potential, ensuring that training remains both challenging and effective. The editor's mutations are designed to allow scenarios to change dynamically at a fine-grained level and can be applied in the following ways:
\begin{itemize}
    \item \textbf{Modify the student’s goal destination}: Alters the student's trajectory, introducing new challenges in decision-making.  
    
    \item \textbf{Adjust the type or attributes of an existing actor}: Modifies an actor class or adjusts its attributes, such as vehicle shape and velocity, to diversify interactions.  

    \item \textbf{Introduce a new actor into an empty node or remove an existing one}: Adds NPCs or obstacles to create more complex traffic or removes them to simplify the scenario, enabling precise control over the scenario.  
\end{itemize}

\subsection{Scenario Curriculum Algorithm}

Our algorithm builds upon ACCEL~\cite{parker2022evolving}, an ACL framework that balances diversity and emerging complexity using a learning potential utility function. This balance reflects the exploration-exploitation trade-off in RL~\cite{sutton1998reinforcement}. Our teacher ensures diversity through a random generator while mutating promising scenarios via an editor. Both teacher components generate driving scenarios by assigning values to free environment parameters  $\boldsymbol{\Theta}$ represented as a directed graph. A key element of the algorithm is the scenario buffer $\Lambda$, which stores scenarios with high learning potential. 

During the algorithm initialization, $\Lambda$ is partially filled to a ratio $\rho \in \left [ 0,1 \right ]$ with randomly generated scenarios, ensuring initial diversity. The ACL algorithm then alternates between sampling unseen scenarios (exploration) and training on a subset of scenarios from $\Lambda$ (exploitation), guided by a replay decision $d$ sampled from a Bernoulli distribution $P_D(d)$ with success probability $D$. 

\subsubsection{\textbf{Exploration Phase}} During exploration ($d=0$), illustrated in Fig. \ref{fig:algorithm_random}, the random generator produces new scenarios using the autoregressive process aforementioned. Each sampled scenario is evaluated based on its learning potential and is only added to the buffer if its learning potential exceeds the current minimum value in $\Lambda$. This approach ensures that $\Lambda$ stores only scenarios that actively contribute to the student’s learning progress.

While previous works~\cite{jiang2021prioritized, jiang2021replay, parker2022evolving} have proposed various approximations of learning potential, we adopt \textbf{positive value loss}~\cite{parker2022evolving}, defined in Eq.~\ref{eq:learning_potential}, as a reliable approximation. The positive value loss is derived from the Generalized Advantage Estimator (GAE)~\cite{schulman2015high}, where $\gamma$ is the discount factor and $\lambda$ is a parameter of GAE. $\delta_t$ is the TD-error at step $t$, as depicted in Eq.~\ref{eq:td_error}, where $V^{\pi}_{\theta}(s_t)$ is value function under a student policy $\pi$ for state $s_t$ in driving scenario with parameters $\theta$.
\begin{equation}
    {U_{\text{pvl}}(\pi, \theta)}= \frac{1}{T}\sum_{t=0}^{T} \, \max \left( \, \sum_{k=t}^{T} (\gamma \lambda)^{k-t} \,\delta_k(\pi, \theta)\,, 0 \,\right)\label{eq:learning_potential}
\end{equation}

\begin{equation}
    \delta_t(\pi, \theta) = r_t + \gamma V^{\pi}_{\theta}(s_{t+1}) - V^{\pi}_{\theta}(s_t) \label{eq:td_error}
\end{equation}

The key advantage of leveraging this learning potential is its role as a policy-aware utility function, adapting to the student's current policy based on errors in value function estimation. This utility function eliminates the need for expert-defined heuristics and remains widely applicable across RL algorithms. Furthermore, positive value loss satisfies a fundamental criterion for effective curriculum generation: a gradual increase in difficulty. By design, it assigns low learning potential to scenarios that are either too simple or overly challenging relative to the student's current abilities, while those that appropriately challenge the student receive high learning potential.

\subsubsection{\textbf{Exploitation Process}} When the decision is to exploit ($d=1$), depicted in Fig.~\ref{fig:algorithm_edit}, we sample a subset of scenarios $\left\{\theta_1,\cdots ,\theta_B \right\}$ from $\Lambda$ to train the student. The probability of sampling a scenario $\theta_i$ is given by $P_{replay}$, a weighted linear combination of the learning potential $P_U$ and the staleness $P_C$ probabilities, with the weight controlled by $\omega$, as shown in Eq.~\ref{eq:replay_probability}. The learning potential probability $P_U$ for a given scenario is proportional to its rank priority relative to all other scenarios in the buffer $\Lambda$~\cite{jiang2021prioritized}. This rank-based prioritization ensures that scenarios with the highest learning potential are sampled more frequently for student training while minimizing the impact of noise in learning potential estimation. On the other hand, the staleness probability $P_C$ is proportional to the number of episodes that have passed since the scenario was last sampled. This probability measure helps prioritize scenarios that have been in $\Lambda$ longer, ensuring their learning potential remains aligned with the current student policy.
\begin{equation} \label{eq:replay_probability}
    P_{replay}(\theta_i) = \omega \, P_U(\theta_i | \Lambda, U) + (1-\omega) \, P_C(\theta_i | \Lambda, C)
\end{equation}

After the student is trained on the sampled subset of scenarios, the editor applies random mutations to these scenarios, with the number of mutations denoted as $N_m$. Inspired by evolutionary algorithms~\cite{slowik2020evolutionary}, the editor assumes that a scenario's learning potential (\textit{fitness}) evolves smoothly through small edits. Thus, starting from a high-learning-potential (\textit{high fitness}) scenario, applying different mutations generates a set (\textit{generation}) of new scenarios with similarly high learning potential. These scenarios might otherwise be difficult to discover through random search due to the vast parameter space.

Once again, to maintain training efficiency, the mutated scenarios are evaluated based on their learning potential and added to $\Lambda$ only if their learning potential exceeds the lowest value in $\Lambda$. The core phase of the algorithm continues until the student converges or the training budget is exhausted. Notably, our proposed algorithm is applicable to any student, regardless of their observation sensors or action space. The only requirement is an estimate of TD-error, a widely available metric in most RL algorithms, to assess learning potential.

\section{experimental Setup}
This section details the experimental setup of this work. We begin by describing the student used in our experiments, followed by an overview of the driving environment for curriculum generation. Finally, we define the evaluation metrics and introduce the baselines for comparison.
\subsection{Student Description}
Our student's observation space consists of RGB images with a $256 \times 256$ resolution. Additionally, the student receives vehicle measurements, including longitudinal and angular velocities, as well as longitudinal and lateral accelerations. A CNN encoder~\cite{mnih2015human} processes the observation space into a latent representation, which is then used to estimate the action distribution and value function. Similar to~\cite{bogdoll2024informed,moghadam2020end}, we use a continuous action space defined by the terminal conditions of a Frenet trajectory $(v_f, d_f)$, where $v_f$ and $d_f$ denote the desired velocity and lateral offset from the centerline at the end of the planning horizon. The student is trained using an on-policy RL algorithm, Proximal Policy Optimization (PPO)~\cite{schulman2017proximal}, across eight parallel simulator instances. As previously mentioned, the observation space, network architecture, and action space can be modified without any constraints imposed by the proposed framework.
\subsection{Driving Environment Setup} 
In this work, we focus on unsignalized intersections, which are common in urban driving and necessitate implicit negotiation between the student and other NPCs. Additionally, these intersections often present safety-critical situations~\cite{khaitan2022state}. The proposed framework is realized in CARLA~\cite{dosovitskiy2017carla}. The graph representation of a diverse set of T-intersections and 4-way intersections is extracted using the CARLA map API, with nodes randomly offset longitudinally and laterally to prevent the student from overfitting to the initial positions of NPCs. To further enhance the student’s visual robustness, the appearances of the NPCs are randomly sampled from Carla's catalog. During evaluation, we assess the agent's performance on intersections from a hold-out set that the student has not encountered during its training.
\subsection{Baselines and Evaluation Metrics}
We evaluate the impact of our proposed framework by comparing it against two commonly used training frameworks for RL agents in AD. The first baseline involves training the student on a fixed set of driving scenarios. The second, DR, trains the student in randomly sampled scenarios without a predefined sequence. To ensure a fair comparison, the student is trained for the same number of steps using an identical architecture and hyperparameters across all frameworks. Table~\ref{tab:params} summarizes the parameters of our proposed algorithm, which were empirically selected to balance learning stability and promote a smooth curriculum. In particular, we set $D$ to $0.8$ encouraging the framework to more frequently modify promising scenarios, enabling the generation of novel scenarios that are suitably challenging for the student. The parameter $\omega$ is set to $0.7$ to bias scenario sampling from $\Lambda$ more toward a scenario’s learning potential than its staleness.

We assess the learned policy using a hold-out set of intersections with varying traffic densities, where density is defined as the ratio of the number of spawned actors (NPCs and obstacles) to the maximum number of actors permitted in a given scenario. The scenarios are randomly generated once and shared across all frameworks to ensure consistency and enable fair comparison. The comparison is based on evaluation metrics such as cumulative reward per episode and standard driving metrics. These include terminal statistics, such as the frequency of successful episodes, off-road incidents, and collisions, as well as driving behavior metrics, such as route progress and average velocity.
\begin{table}[t]
\centering
\renewcommand{\arraystretch}{1.3}
\caption{The parameters of the proposed ACL algorithm.}
\resizebox{0.7\columnwidth}{!}{
\begin{tabular}{cc|cc}
\toprule
\textbf{Parameter} & \textbf{Value} & \textbf{Parameter} & \textbf{Value}\\\midrule
$N$       & 1000  & $\lambda$ & 0.9   \\
$\rho$    & 0.5   & $B$         & 32    \\
$D$  & 0.8   & $\omega$  & 0.7   \\
$\gamma$  & 0.99  & $N_m$     & 2  \\ \bottomrule
\end{tabular}
}
\label{tab:params}
\vspace{-0.2cm}
\end{table}

\section{Evaluation}
In this section, we present the results of our proposed framework. First, we evaluate its ability to generate a smooth curriculum of driving scenarios. Next, we compare the performance of agents trained with our framework against baseline methods and analyze its impact on training efficiency.
\subsection{Scenario Curriculum Generation}
\begin{figure}[t]
    \centering
    \includegraphics[width=0.8\linewidth]{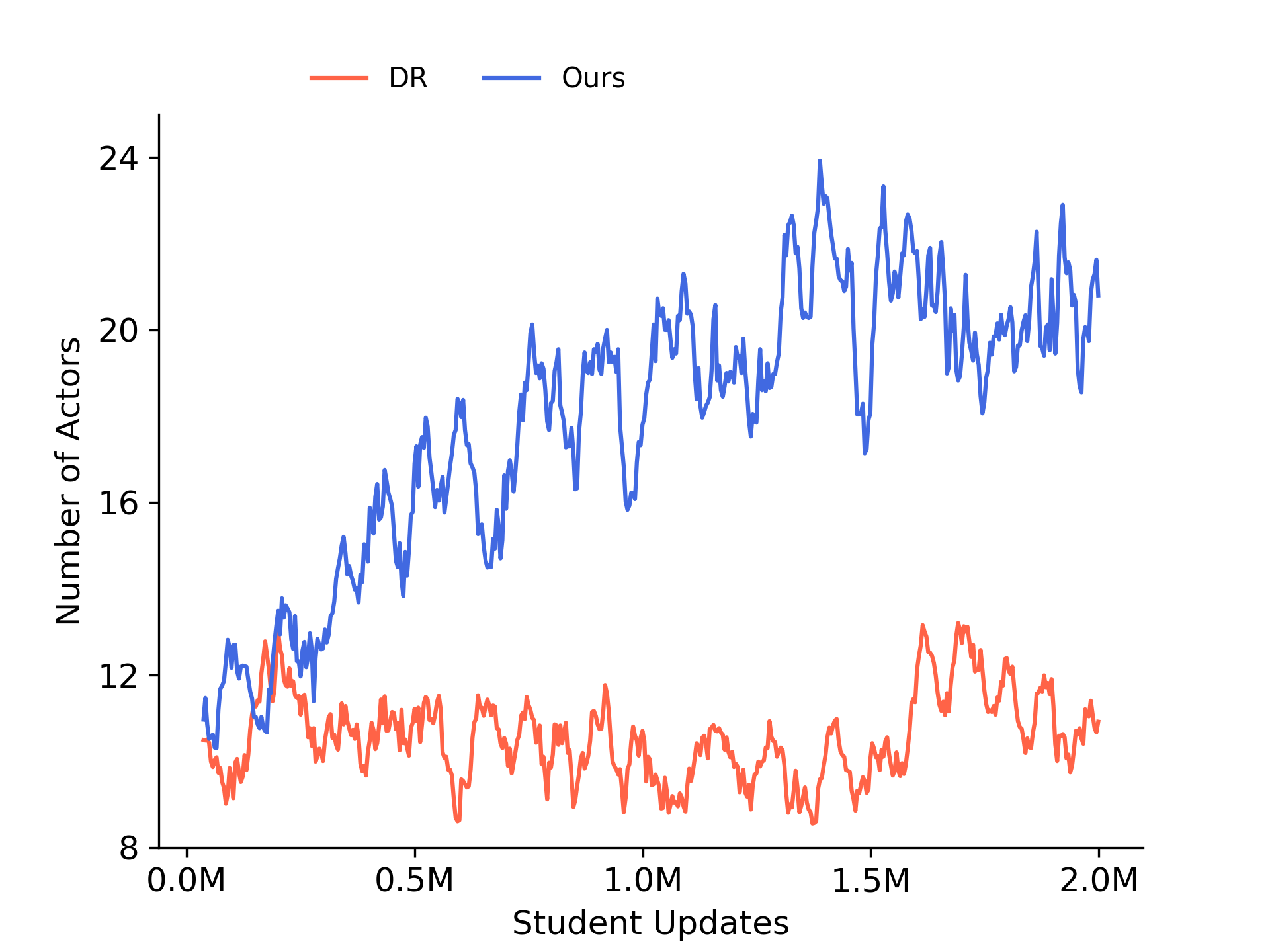}
    \caption{A comparison between the average number of actors in the training scenarios generated by DR (red) and the proposed ACL framework (blue). The figure illustrates that our framework can generate a curriculum with an evolving complexity compared to random scenarios from DR.}
    \label{fig:number_of_actors_vs_steps}
    \vspace{-0.375cm}
\end{figure}

The goal of curriculum learning is to create training tasks (driving scenarios) with progressively increasing difficulty. A smooth curriculum can reduce the high variance in training data observed in DR. Fig.~\ref{fig:number_of_actors_vs_steps} illustrates the number of actors (NPCs and obstacles) in the subset of training scenarios sampled per student update, averaged over all parallel simulator instances. This provides insight into the quality of the curriculum learned by the proposed framework. Our framework demonstrates a gradual increase in the number of actors as the student interacts with the environment over time, in contrast to DR, which exhibits high variance due to random sampling. 

\begin{figure*}[t!]
    \begin{subfigure}[t]{0.325\linewidth}
        \begin{subfigure}[t]{0.49\linewidth}
            \centering
            \includegraphics[width=\linewidth]{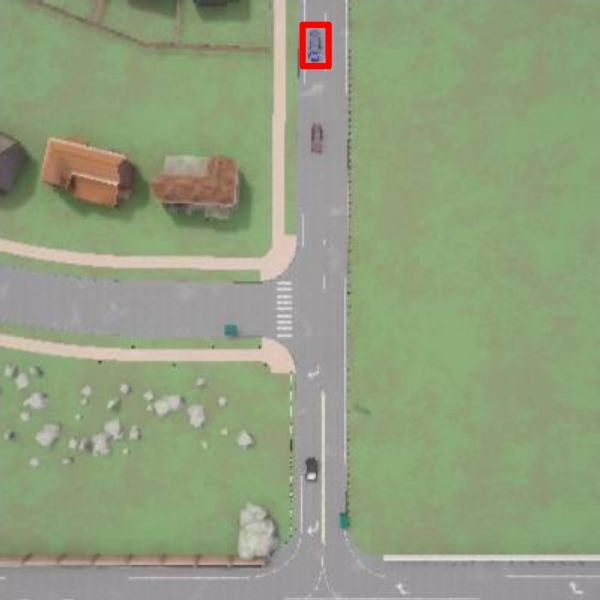}
        \end{subfigure}%
        \hfill
        \begin{subfigure}[t]{0.49\linewidth}
            \centering
            \includegraphics[width=\linewidth]{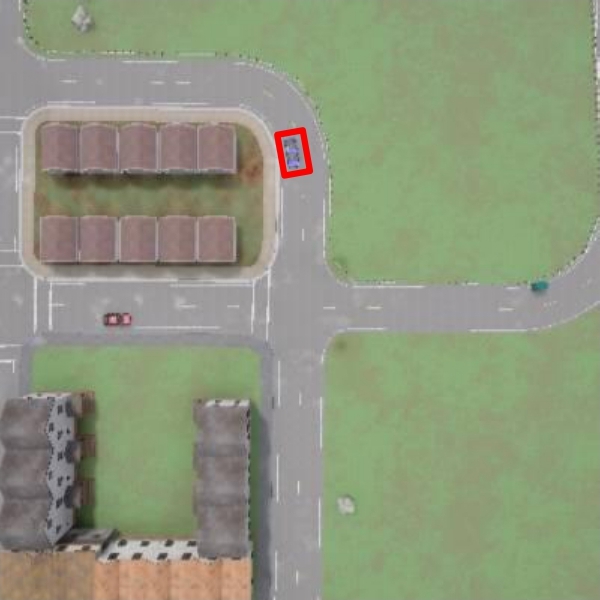}
        \end{subfigure}%
        \caption{Exemplary scenarios generated at the beginning of the student training.}
    \end{subfigure}%
    \hfill
    \begin{subfigure}[t]{0.325\linewidth} 
        \begin{subfigure}[t]{0.49\linewidth}
            \centering
            \includegraphics[width=\linewidth]{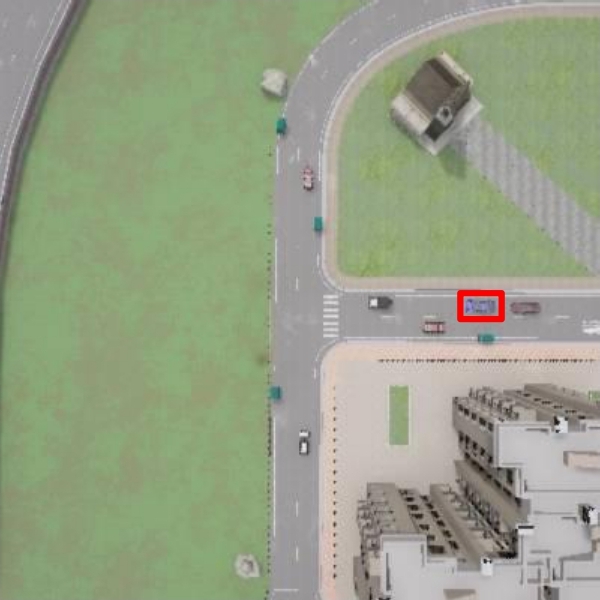}
        \end{subfigure}%
        \hfill
        \begin{subfigure}[t]{0.49\linewidth}
            \centering
            \includegraphics[width=\linewidth]{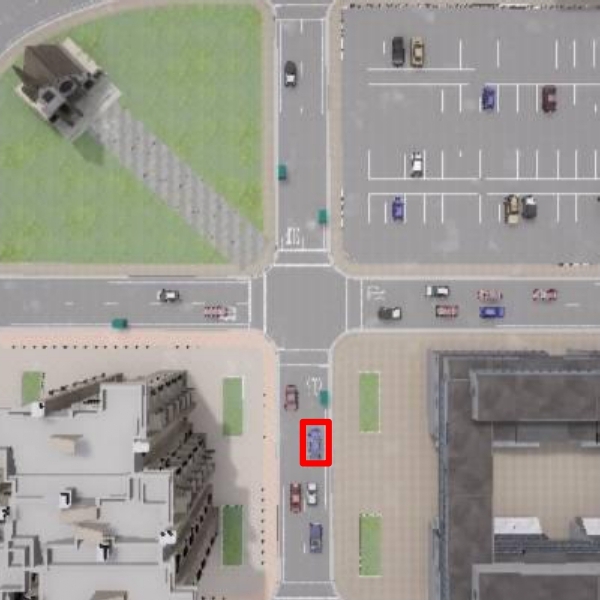}
        \end{subfigure}%
        \caption{Exemplary scenarios generated during the intermediate phase of the student training.}
    \end{subfigure}%
    \hfill
    \begin{subfigure}[t]{0.325\linewidth} 
    \centering
        \begin{subfigure}[t]{0.49\linewidth}
            \centering
            \includegraphics[width=\linewidth]{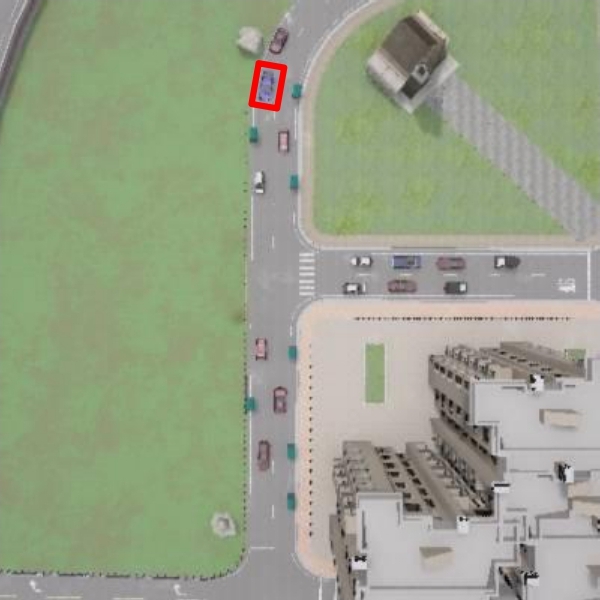}
        \end{subfigure}%
        \hfill
        \begin{subfigure}[t]{0.49\linewidth}
            \centering
            \includegraphics[width=\linewidth]{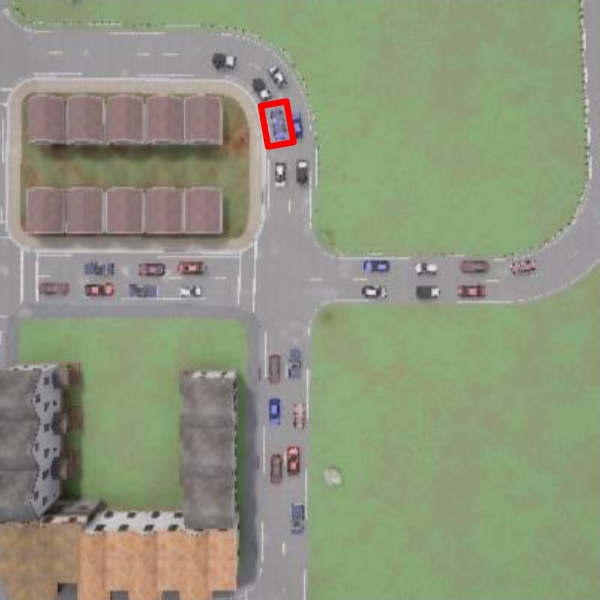}
        \end{subfigure}%
        \caption{Exemplary scenarios generated at the final phase of the student training.}
    \end{subfigure}
    \caption{Exemplary scenarios generated by the proposed ACL framework at different phases of the student training. It can be observed that the scenarios have emerging complexity as the training progresses.}
    \label{fig:interaction_types}
\end{figure*}

Notably, certain updates show a decrease in the number of actors, indicating that scenarios are dynamically adjusted to the student's current capabilities rather than increasing complexity uniformly. As training progresses, our framework generates increasingly challenging scenarios with a higher number of actors compared to DR, underscoring its ability to evolve training conditions. Additionally, Fig.~\ref{fig:interaction_types} presents examples of scenarios generated by our framework at various training phases, illustrating the evolving complexity and their realization in CARLA.
\subsection{Policy Evaluation}
We further evaluate the impact of training an RL agent within the proposed scenario curriculum. Table~\ref{table:final_metrics} presents a comparative analysis of the student policy’s performance after training, contrasting our framework with the baselines across various performance metrics. For continuous metrics, both the mean and standard deviation are reported to provide a comprehensive evaluation. The student trained within the proposed framework demonstrates a higher success percentage and fewer collisions, increasing success by \textbf{9\% in low traffic density} and \textbf{21\% in high traffic density} in the hold-out set of intersection environments. Additionally, the student consistently outperforms the baselines in maximizing cumulative reward and route progress, while exhibiting lower standard deviation, indicating more consistent performance across different scenarios. 

Although training in fixed environments results in velocities closer to the desired value of 3 $m/s$, this comes at the cost of a higher collision rate. In contrast, our framework enables higher velocities while maintaining fewer collisions than DR. These findings indicate that the student policy not only improves performance but also generalizes more effectively to complex tasks, such as navigating unsignalized intersections with varying numbers of actors moving at different velocities and offsets.
\begin{table*}[h]
\renewcommand{\arraystretch}{1.3}
\centering
\caption{The evaluation metrics assessed on the student’s final policy trained with different frameworks (\textit{fixed, DR, Ours}), using a hold-out set of intersection environments. These environments were randomly generated with traffic density varying from \textit{0.5} to \textit{1.0}.}
\resizebox{\linewidth}{!}{
\begin{tabular}{l|rrr|rrr|rrr}
\toprule
& \multicolumn{3}{c|}{Traffic Density 0.5}        & \multicolumn{3}{c|}{Traffic Density 0.75}                         & \multicolumn{3}{c}{Traffic Density 1.0} \\ \cline{2-4}\cline{5-7}\cline{8-10}
Metrics                                        & Fixed                  & DR              & Ours                       & Fixed                  & DR              & Ours                     & Fixed                  & DR              & Ours                     \\ \midrule
Success (\%) $\boldsymbol{\uparrow}$           & 45.0                   & 71.0            & \textbf{80.0}              & 31.0                   & 55.0            & \textbf{68.0}            & 16.0                   & 35.0            & \textbf{56.0}            \\
Off-road (\%) $\boldsymbol{\downarrow}$        & 2.0                    & 0.0             & 0.0                        & 2.0                    & 0.0             & 0.0                      & 1.0                    & 0.0             & 0.0                      \\
Collision (\%) $\boldsymbol{\downarrow}$       & 53.0                   & 29.0            & \textbf{20.0}              & 67.0                   & 45.0            & \textbf{32.0}            & 83.0                   & 65.0            & \textbf{44.0}            \\ \midrule
Cumulative Reward $\boldsymbol{\uparrow}$      & $24.59 \pm 50.14$      & $49.13\pm44.08$ & $\mathbf{63.77} \pm 38.43$ & $8.18\pm46.35$         & $36.35\pm46.73$ & $\mathbf{51.29}\pm45.84$ & $-6.34\pm36.05$        & $17.00\pm45.49$ & $\mathbf{37.83}\pm49.16$ \\ \midrule
Route Progress $\boldsymbol{\uparrow}$         & $0.62\pm0.35$          & $0.75\pm0.32$   & $\mathbf{0.83}\pm0.22$     & $0.48\pm0.35$          & $0.69\pm0.31$   & $\mathbf{0.76}\pm0.28$   & $0.38\pm0.33$          & $0.54\pm0.35$   & $\mathbf{0.65}\pm 0.35$  \\
Average Velocity (m/s) $\boldsymbol{\uparrow}$ & $\mathbf{3.14}\pm0.52$ & $2.18\pm0.21$   & $2.34\pm0.12$              & $\mathbf{3.09}\pm0.55$ & $2.12\pm0.21$   & $2.33\pm0.14$            & $\mathbf{2.94}\pm0.59$ & $1.95\pm0.25$   & $2.27\pm0.18$            \\ \bottomrule
\end{tabular}
}
\label{table:final_metrics}
\end{table*}
\subsection{Training Efficiency}
In this section, we evaluate whether the student can achieve better performance on the hold-out set within fewer updates compared to the baselines. We analyze the success rate and cumulative reward obtained by the student throughout training, comparing DR and the proposed framework, as shown in Tables~\ref{table:checkpoint_success}, \ref{table:checkpoint_reward}. The comparison primarily focuses on DR, as it has the closest performance to the proposed framework. We begin the evaluation after one million student updates, when exploration diminishes, indicated by the distribution entropy falling below zero. The results clearly demonstrate that the proposed framework consistently outperforms DR, achieving higher cumulative rewards and success rates at all training phases and across varying traffic densities. This supports our hypothesis that a curriculum of driving scenarios, adaptively generated based on the student's current capabilities, enhances training efficiency and leads to faster convergence.
\begin{table}[]
\centering
\caption{The \textit{success percentage} ($\boldsymbol{\uparrow}$) assessed across different phases of the training (\textit{Student Updates}) with different frameworks (\textit{DR, Ours}), using a hold-out set of intersection environments. These environments were randomly generated with traffic densities of \textit{0.75} and \textit{1.0} respectively}
\renewcommand{\arraystretch}{1.3}
\resizebox{0.4\textwidth}{!}{
\begin{tabular}{l|rr|rr}
\toprule & \multicolumn{2}{c|}{Traffic Density 0.75}          & \multicolumn{2}{c}{Traffic Density 1.0} \\  \cline{2-3}\cline{4-5}
Student Updates & DR & Ours & DR & Ours \\ \midrule
1.00 M          & 25.0                   & \textbf{58.0}             & 12.0                   & \textbf{50.0}            \\
1.25 M          & 14.0                   & \textbf{57.0}             & 4.0                    & \textbf{37.0}            \\
1.50 M          & 39.0                   & \textbf{65.0}             & 29.0                   & \textbf{51.0}            \\
1.75 M          & 43.0                   & \textbf{66.0}             & 38.0                   & \textbf{51.0}            \\
2.00 M          & 55.0                   & \textbf{68.0}             & 35.0                   & \textbf{56.0}            \\ \bottomrule
\end{tabular}
}
\label{table:checkpoint_success}
\end{table}
\begin{table}[h]
\caption{The \textit{cumulative reward} ($\boldsymbol{\uparrow}$) assessed across different phases of the training (\textit{Student Updates}) with different frameworks (\textit{DR, Ours}), using a hold-out set of intersection environments. These environments were randomly generated with traffic densities of \textit{0.75} and \textit{1.0} respectively}
\renewcommand{\arraystretch}{1.3}
\resizebox{0.48\textwidth}{!}{
\begin{tabular}{l|rr|rr}
\toprule
& \multicolumn{2}{c|}{Traffic Density 0.75}    & \multicolumn{2}{c}{Traffic Density 1.0}  \\  \cline{2-3}\cline{4-5}
Student Updates & \multicolumn{1}{c}{DR} & \multicolumn{1}{c|}{Ours}  & \multicolumn{1}{c}{DR} & \multicolumn{1}{c}{Ours}   \\ \midrule
1.00 M          & $1.29 \pm 31.78$  & $\mathbf{40.16} \pm 48.03$ & $-11.35 \pm 23.99$ & $\mathbf{32.05} \pm 49.05$ \\
1.25 M          & $-5.96 \pm 24.66$ & $\mathbf{40.44} \pm 47.70$ & $-16.07 \pm 15.55$ & $\mathbf{20.47} \pm 46.84$ \\
1.50 M          & $15.37 \pm 39.55$ & $\mathbf{47.54} \pm 45.95$ & $4.26 \pm 34.58$   & $\mathbf{31.90} \pm 49.46$ \\
1.75 M          & $17.28 \pm 37.85$ & $\mathbf{48.89} \pm 46.75$ & $12.44 \pm 38.41$  & $\mathbf{35.14} \pm 49.55$ \\
2.00 M          & $36.35 \pm 46.73$ & $\mathbf{51.29} \pm 45.84$ & $17.00 \pm 45.49$  & $\mathbf{37.83} \pm 49.16$ \\ \bottomrule
\end{tabular}
}
\vspace{-0.5cm}
\label{table:checkpoint_reward}
\end{table}

\section{Conclusion}

In this paper, we presented a novel ACL framework that enhances the training efficiency and generalization capabilities of RL-based E2E driving agents. Our approach leverages a graph-based representation of driving environments and employs a teacher component that generates and mutates driving scenarios based on their learning potential. This method overcomes the limitations of fixed training and DR by providing a structured and scalable curriculum that adapts to the student’s capabilities. Our experimental results demonstrate that the proposed framework outperforms the baselines in terms of training efficiency and policy generalization, achieving higher success rates and fewer collisions in unseen scenarios. These findings underscore the potential of ACL in developing more robust E2E driving agents. Future work will focus on incorporating non-road-bound NPCs, such as pedestrians and cyclists, to further increase the complexity of training scenarios. Additionally, we plan to explore advanced scenario mutation techniques, including the use of deep learning for scenario editing.

{
    \bibliographystyle{IEEEtran}
    
    \bibliography{references}
}

\end{document}